\documentclass[nohyperref]{article}

\usepackage{microtype}
\usepackage{graphicx}
\usepackage{subfigure}
\usepackage{booktabs} 
\usepackage{multirow}

\usepackage{hyperref}

\usepackage[accepted]{icml2022}

\usepackage{amsmath}
\usepackage{amssymb}
\usepackage{mathtools}
\usepackage{amsthm}
\usepackage[capitalize,noabbrev]{cleveref}

\theoremstyle{plain}

\theoremstyle{definition}

\theoremstyle{remark}

\usepackage[textsize=tiny]{todonotes}

\icmltitlerunning{Efficient Fine-Tuning with Learners}

\begin{document}

\twocolumn[
\icmltitle{Efficient Fine-Tuning of Compressed Language Models with Learners}




\begin{icmlauthorlist}
\icmlauthor{Danilo Vucetic}{mcg}
\icmlauthor{Mohammadreza Tayaranian}{mcg}
\icmlauthor{Maryam Ziaeefard}{mcg}
\icmlauthor{James J. Clark}{mcg}
\icmlauthor{Brett H. Meyer}{mcg}
\icmlauthor{Warren J. Gross}{mcg}
\end{icmlauthorlist}

\icmlaffiliation{mcg}{Department of Electrical and Computer Engineering, McGill University, Montreal, Canada}

\icmlcorrespondingauthor{Danilo Vucetic}{danilo.vucetic@mail.mcgill.ca}

\icmlkeywords{Machine Learning, ICML, BERT, DistilBERT, Efficient Fine-Tuning, Fine-Tuning, Transfer Learning, NLP, Natural Language Processing, Efficient Machine Learning, Efficient Training, Language Modelling, Learners, Learner Module, Adapter, Adapter Module}

\vskip 0.3in
]



\printAffiliationsAndNotice{}  

\begin{abstract}
Fine-tuning BERT-based models is resource-intensive in memory, computation, and time. While many prior works aim to improve inference efficiency via compression techniques, e.g., pruning, these works do not explicitly address the computational challenges of training to downstream tasks. We introduce \textit{Learner} modules and priming, novel methods for fine-tuning that exploit the overparameterization of pre-trained language models to gain benefits in convergence speed and resource utilization. Learner modules navigate the double bind of 1) training efficiently by fine-tuning a subset of parameters, and 2) training effectively by ensuring quick convergence and high metric scores. Our results on DistilBERT demonstrate that learners perform on par with or surpass the baselines. Learners train 7x fewer parameters than state-of-the-art methods on GLUE. On CoLA, learners fine-tune 20\% faster, and have significantly lower resource utilization. 

\end{abstract}

\section{Introduction}
Transformer-based Pre-trained Language Models (PLM) have become ubiquitous in Natural Language Processing (NLP). BERT and its various derivatives outperform the previous generations of NLP models significantly, requiring in contrast many more parameters and, consequently, more powerful hardware resources for training and inference \cite{energy_in_NLP, green_ai, devlin2018bert}. Works in model compression have sought to improve inference efficiency of these models by pruning, quantization, and distillation \cite{Ganesh2021CompressingLT, survey_on_device_ML}. Fine-tuning is, however, complicated by the size of PLMs \cite{energy_in_NLP, green_ai}. For example, BERT and DistilBERT have parameter counts of 110 million and 66 million parameters respectively \cite{devlin2018bert, distilbert}. Fine-tuning such large models requires substantial data and considerably more computations and memory accesses than inference \cite{tinytl, green_ai, energy_in_NLP}. Further, memory operations are energy intensive and slow \cite{songhan_eie, comp_arch_hennessy_patterson_2012}. Fine-tuning must be made fast and resource-efficient to facilitate use-cases such as personalized auto-correct systems on mobile device keyboards, or green AI \cite{federated_mobile_keyboard_example, survey_on_device_ML, green_ai}.



Prior works in efficient training focus primarily on parameter efficiency for uncompressed (i.e., huge) transformer-based models \cite{bitfit, houlsby2019parameter_eff_adapter, he2022towards_unified_param_eff_parallel_adapter, guo-etal-2021-parameter-diff-pruning}. Adapters \cite{houlsby2019parameter_eff_adapter} manage to achieve parameter efficiency in a computationally efficient manner, while difference pruning \cite{guo-etal-2021-parameter-diff-pruning} requires triple the training-time parameter usage. We show that the computational efficiency of adapters is undermined on a compressed model by slow convergence. State-of-the-art methods in efficient fine-tuning, namely Freeze-and-Reconfigure (FAR), are decidedly quick to converge, but require a far greater proportion of parameters and are plagued by slow memory operations \cite{FAR}. The optimal efficient fine-tuning technique must then satisfy a few requirements: 1) demonstrate quick convergence, and 2) be resource efficient in memory, computation, and time.

\begin{figure*}[ht]
\begin{center}
\includegraphics[width=0.9\textwidth]{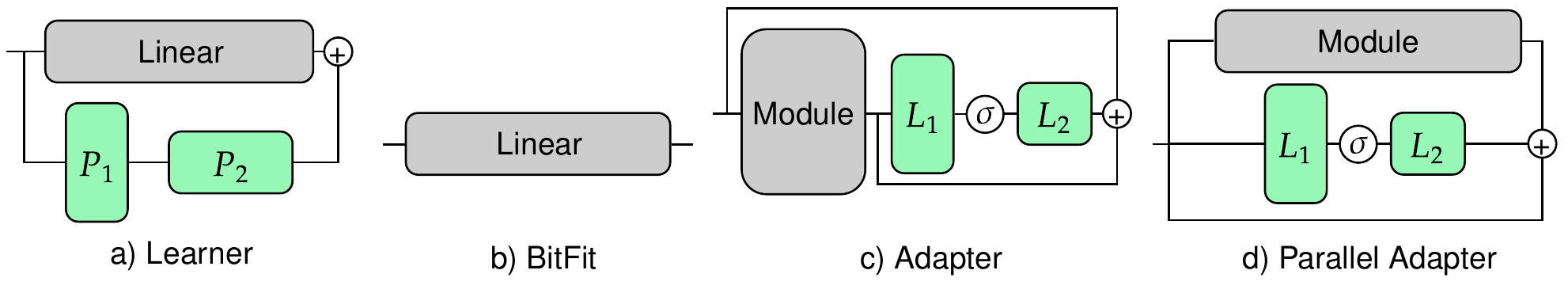}
\vskip -0.1in
\caption{Comparison of learner modules against other architectures. Grey indicates a frozen (i.e., not trained) component, green indicates trained components. Our proposed modules, learners, are shown in a) where a linear layer is frozen except for the bias and the two projection matrices, $P_1 \text{and} P_2$, are trained. BitFit is illustrated in b) where the linear layer is frozen except for the bias. Adapters and parallel adapters are shown in c) and d). Both have the same architecture but different placements. The adapter consists of two linear layers, $L_1 \text{and} L_2$, separated by an activation function with a residual connection, similar to the FFN in a BERT model.}
\label{figure_compare_learner}
\end{center}
\vskip -0.2in
\end{figure*}

Our main contribution is the proposal of \textit{Learner} modules fine-tuned with priming steps, which satisfy these requirements by: a) exploiting pre-trained parameters for their quick convergence, b) exploiting model overparameterization to train a small subset of parameters, and c) avoiding slow, complex memory operations. Learner modules, as illustrated in Figure \ref{figure_compare_learner}, are added in parallel to each linear layer in a target model. Learners consist of a low-rank projection matrix whose product with module input is added to the output of the linear layer. In essence, the small learner module learns for the much larger linear layer. After training, the learner modules can be collapsed leaving just the original model, unlike adapters which permanently add modules to the architecture. In this work we demonstrate the importance of convergence for efficient fine-tuning. We show that learner modules outperform adapters by 3 GLUE points while adding a similar number of parameters, because of their quicker convergence. We also demonstrate that learners perform on par with FAR while training 7x fewer parameters, and fine-tuning 20\% faster.

\section{Background and Related Work}

Parameter-efficient fine-tuning techniques attempt to reduce the number of parameters per fine-tuning task. This is done to reduce the cost of storing multiple sets of parameters on, for example, a server which executes multiple tasks based on client requests. Parameter-efficient techniques may also be computationally efficient. Adapter modules are in essence small BERT-like Feed-Forward Networks (FFNs) which are exclusively trained alongside layer normalization parameters \cite{houlsby2019parameter_eff_adapter}. Adapters may be placed sequentially with modules or in parallel \cite{he2022towards_unified_param_eff_parallel_adapter}. Due to their small trained parameterization (i.e., the proportion of trained parameters), adapters are computationally efficient to train. The same goes for a number of methods such as BitFit, which trains only the biases of BERT models while maintaining good metric performance \cite{bitfit}. Difference pruning is decidedly not computationally efficient, requiring three times the parameters of a given model to achieve a parameter-efficient fine-tuning result. Difference pruning seeks to prune not the model itself, but the learned values added to pre-trained parameters during fine-tuning. In doing so, a tiny fraction of parameters are trained while producing excellent metric results \cite{guo-etal-2021-parameter-diff-pruning}. In the review process we were made aware of a Low-Rank Adaptation (LoRA) method by Hu et al. \cite{hu2021lora}, which uses a similar architecture to Learners, but with a scaling factor after the second projection matrix. The difference between LoRA and Learners is mainly that Learners employ a priming step, which is shown to improve convergence significantly on a compressed model. All parameter-efficient methods relevant to efficient fine-tuning are illustrated in Figure \ref{figure_compare_learner}. The commonality of parameter-efficient methods is their exploration only in large transformer-based models (all are trained on $\text{BERT}_\text{LARGE}$ or similarly large models). A question remains in their efficacy on compressed models: Can parameter-efficient methods efficiently train compressed models that lack the parameterization of their proposed models?

Efficient fine-tuning attempts to reduce the computational costs of fine-tuning. Freeze-and-Reconfigure (FAR) efficiently fine-tunes DistilBERT by strategically freezing subsets of parameters in the FFNs that do not learn quickly after an initial number of training iterations, called priming steps \cite{FAR}. The FFNs are reconfigured to explicitly separate these quick-learning parameters from the slower ones. While FAR is effective at reducing the trained parameterization, training time, and memory access time, it still requires memory-intensive permutations to coherently combine results of FFN computations. FAR also requires substantially more parameters than adapters (40 million in DistilBERT versus 5 million in $\text{BERT}_\text{LARGE}$ respectively). BitFit and the various adapter configurations and architectures have not yet been tested on compressed language models. They are certainly efficient to fine-tune in terms of computational cost, but their metric performance, convergence speed, and training time must be considered.

\section{The Necessaries of Efficient Fine-Tuning}

\begin{table*}[!htbp]
\caption{Metrics of different efficient fine-tuning methods training on CoLA and running on an NVIDIA Jetson Xavier NX for 20 epochs. Total parameters are the full training-time size of the model. Training time is measured as wall-clock time. Peak memory is measured using CUDA calls in PyTorch, which report the maximum amount of memory allocated during fine-tuning. Learners are presented fully later, but note that their parameterization changes over training, starting off from around 20 million and dropping after priming to 5.96 million. Peak memory in later epochs is thus much lower, but not reflected by this overall measure.}
\label{table_paramerization}
\small
\centering
\addtolength{\tabcolsep}{3pt}
\begin{tabular}{@{}lcccc@{}}
\toprule
Method &
  \begin{tabular}[c]{@{}c@{}}Total Parameters\\ ($\times 10^6$)\end{tabular} &
  \begin{tabular}[c]{@{}c@{}}Trained Parameters\\ ($\times 10^6$)\end{tabular} &
  \begin{tabular}[c]{@{}c@{}}Time \\ (min)\end{tabular} &
  \begin{tabular}[c]{@{}c@{}}Peak Memory \\ (GB)\end{tabular} \\ \midrule
Baseline                              & 66.95 & 66.95 & 56.87 & 1.306 \\
$\text{FAR}_{10}$                     & 66.95 & 41.45 & 46.60 & 1.017 \\
Adapter                               & 71.68 & 5.34  & 33.01 & 0.554 \\
Parallel Adapter                      & 71.68 & 5.33  & 31.02 & 0.546 \\
\textsc{Freeze FFNs} & 66.95 & 14.80 & 32.59 & 0.630 \\
BitFit                                & 66.95 & 0.64  & 27.73 & 0.440 \\
$\text{Learner}_{64} (p=2)$           & 72.26 & 5.96  & 40.50 & 0.793 \\ \bottomrule
\end{tabular}
\end{table*}

\begin{figure}[ht]
\begin{center}
\centerline{\includegraphics[width=0.88\columnwidth]{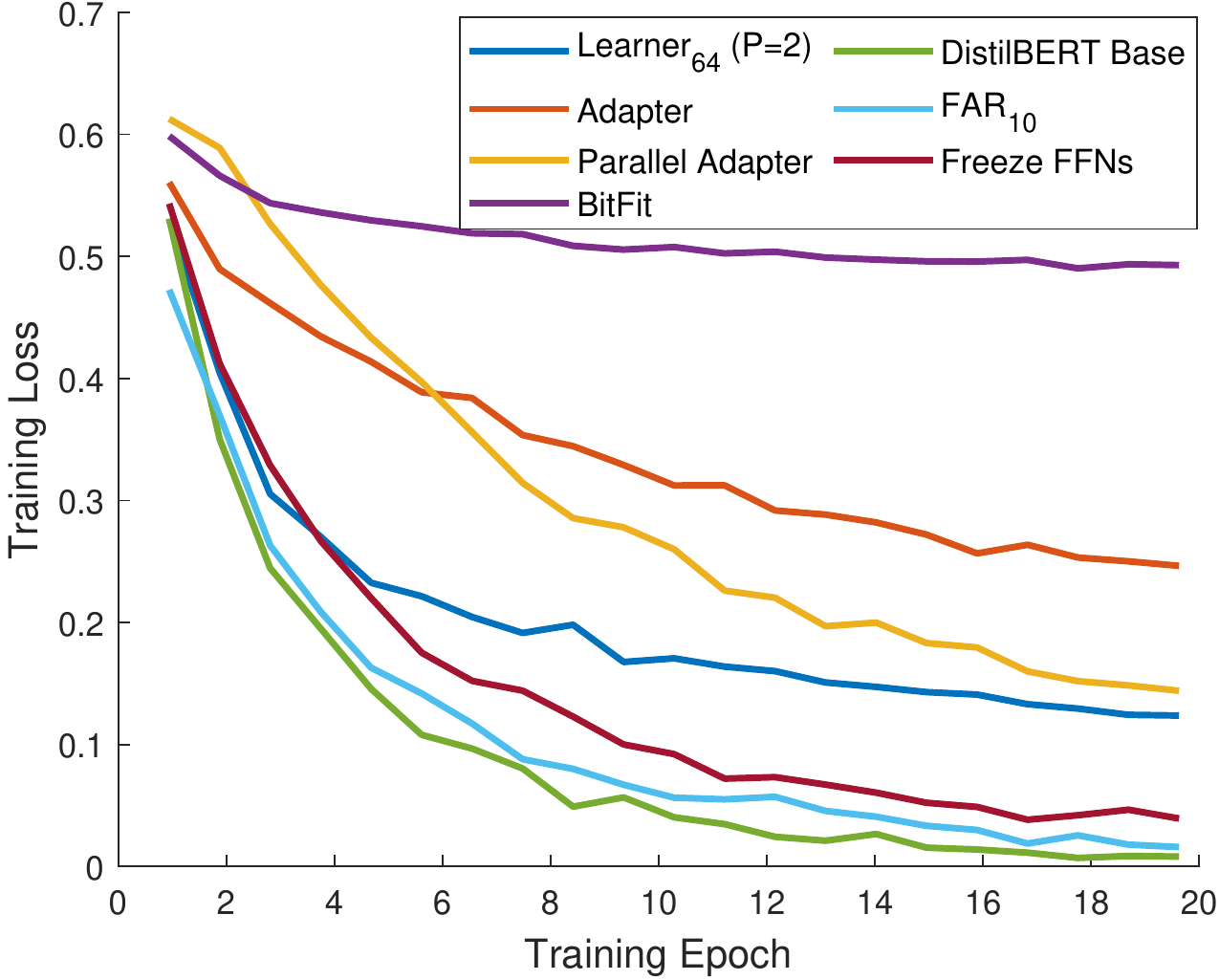}}
\caption{Convergence of training loss of DistilBERT on CoLA with different fine-tuning methods. Adapters and BitFit clearly struggle to converge in 20 epochs while $\text{FAR}_{10}$ and other methods that directly train the pre-trained parameters converge quickly and to lower loss values. \textsc{Freeze FFNs} trains only the multi-head attention of DistilBERT. Learners, which are presented later, converge at a rate between the two. These trends hold for all GLUE tasks as demonstrated in the appendix.}
\label{figure_cola_convergence}
\end{center}
\vskip -0.3in
\end{figure}

Efficient fine-tuning methods must navigate the double bind of 1) training efficiently by reducing the number of trained parameters thereby reducing resource utilization (i.e., time, memory, compute), and 2) training effectively by ensuring quick convergence and high training scores (i.e., high GLUE scores, SQuAD accuracy, etc.). Methods that add parameters to a model, such as adapters, are still able to train efficiently since only the added parameters are trained alongside negligible layer normalization parameters \cite{houlsby2019parameter_eff_adapter}. However, adapters and related methods tend to train smaller datasets such as CoLA for 20 epochs while FAR can achieve on par performance with the baseline after only 5 epochs across all datasets \cite{bitfit, houlsby2019parameter_eff_adapter, he2022towards_unified_param_eff_parallel_adapter, FAR}. When attempting efficient training, it is not enough to just reduce the computational costs. The time to train (a product of the number of epochs and resource usage) must also be considered to enable on-device learning and green AI \cite{green_ai}.

We found that the differential in performance between FAR and parameter-efficient methods may be explained with the convergence of training loss. Figure \ref{figure_cola_convergence} illustrates the training loss of DistilBERT over 20 epochs on the CoLA dataset and Table \ref{table_paramerization} lists relevant performance metrics. It is clear that those methods with a small number of trained parameters tend to take longer to converge and do not converge to the same low-loss level as the baseline. For instance, BitFit converges poorly while adapters require more epochs to converge to a higher loss than baselines. In contrast, the \textsc{Freeze FFNs} curve demonstrates that by training only the multi-head attention, quick convergence can be achieved and to a low loss level while training only 22\% of the model. This may be the reason behind FAR's quick convergence, since it trains the multi-head attention alongside quick-learning FFN nodes. Evidently, fine-tuning pre-trained parameters is important for achieving quick convergence to a low loss, which is required for effective training. However, Table \ref{table_paramerization} also makes clear that FAR and \textsc{Freeze FFNs} train more parameters than adapters. FAR is also more memory intensive than adapters in peak memory. Clearly, to achieve efficient training, complex memory operations and the number of trained parameters must be reduced from these baselines.

\section{Learners}
We propose Learner modules to ensure quick convergence to a low loss value (i.e., achieving a good metric score quickly) and to dynamically reduce the proportion of trained parameters, leading to lower resource utilization. Learner modules are similar to adapters in that they add newly initialized parameters to a model and train these parameters in lieu of the pre-trained model parameters. However, the architectures of these methods differ by both their placements and their components. The Learner module architecture is illustrated in Figure \ref{figure_compare_learner} a). Learners modules consist of two projection matrices added in parallel to every linear layer of a model. Learner modules also do not employ a non-linearity, meaning that they may be collapsed after fine-tuning, unlike adapters, which are permanently fixed to the model. This means that Learner modules do not affect inference efficiency. In addition, the simplicity of their architecture means that Learner modules do not incur any unnecessary overhead in contrast to FAR, whose computationally complex permutations significantly slow down fine-tuning \cite{FAR}. Learner modules by their simple architecture will therefore not impede efficient training.

\begin{table*}[t!]
\addtolength{\tabcolsep}{-3pt}
\caption{Results of all methods on the validation sets of GLUE with averaged GLUE scores across all metrics without WNLI. Datasets reported with training set size and their metric. MCC is Matthew's Correlation, Acc. is accuracy, PC is Pearson Correlation, and SP is Spearman Correlation. MNLI results report both the matched (m) and mismatched (mm) validation sets. Each data point is the average of five experiments with different seeds, each run for five epochs. When Learners have p=0, they are equivalent to LoRA with a scaling factor of 1.}
\label{table_glue_results}
\scriptsize
\begin{tabular}{@{}lccccccccc|r@{}}
\toprule
\multicolumn{1}{c}{\multirow{2}{*}{\textbf{Method}}} & \multicolumn{1}{c}{\textbf{CoLA}} & \multicolumn{1}{c}{\textbf{MNLI m/mm}} & \multicolumn{1}{c}{\textbf{MRPC}}    & \multicolumn{1}{c}{\textbf{QNLI}} & \multicolumn{1}{c}{\textbf{QQP}}     & \multicolumn{1}{c}{\textbf{RTE}} & \multicolumn{1}{c}{\textbf{SST-2}} & \multicolumn{1}{c}{\textbf{STS-B}} & \multicolumn{1}{c|}{\textbf{WNLI}} & \multicolumn{1}{c}{\multirow{2}{*}{\textbf{GLUE Score}}} \\
\multicolumn{1}{c}{}                                 & \multicolumn{1}{l}{MCC - 8.5k}    & \multicolumn{1}{l}{Acc. - 393k}        & \multicolumn{1}{l}{Acc. / F1 - 3.7k} & \multicolumn{1}{l}{Acc. - 105k}   & \multicolumn{1}{l}{Acc. / F1 - 364k} & \multicolumn{1}{l}{Acc. - 2.5k}  & \multicolumn{1}{l}{Acc. - 67k}     & \multicolumn{1}{l}{PC / SC - 7k}   & \multicolumn{1}{l|}{Acc. - 634}    & \multicolumn{1}{c}{}                                     \\ \midrule
Baseline                                             & \textbf{52.56}                    & 82.06 / 82.25                            & \textbf{84.80 / 89.43}                 & 88.39                             & \textbf{90.37 / 87.13}                 & 60.14                            & 90.27                              & 86.85 / 86.52                        & 35.21                              & 79.51                                                    \\
$\text{FAR}_{10}$                                                  & 51.67                             & 81.65 / 81.82                            & 84.85 /  89.48                           & 87.92                             & 90.17 / 86.86                          & \textbf{63.47}                   & \textbf{90.55}                     & 86.57 / 86.29                        & 35.49                              & \textbf{79.68}                                           \\
BitFit                                               & 28.77                             & 68.46 / 70.04                            & 70.98 / 82.45                          & 79.51                             & 82.41 / 77.23                          & 58.34                            & 86.08                              & 47.71 / 45.82                        & 45.92                              & 65.66                                                    \\
Sequential Adapters                                  & 43.24                             & 80.08 / 80.70                            & 80.00 / 86.06                          & 85.67                             & 88.01 / 84.02                          & 56.75                            & 89.88                              & 83.97 / 83.73                        & 41.13                              & 76.10                                                   \\
Parallel Adapters                                    & 18.16                             & 78.09 / 78.97                            & 74.85 / 83.19                          & 85.75                             & 88.50 / 84.76                          & 54.58                            & 89.13                              & 83.86 / 83.59                        & \textbf{51.83}                     & 71.94                                                   \\
\textsc{Freeze FFNs}                                            & 51.49                             & 81.68 / 82.10                            & 84.12 / 88.90                          & 88.68                             & 89.82 / 86.44                          & 60.58                            & 90.30                              & 86.52 / 86.22                        & 37.75                              & 79.24                                                   \\
$\text{Learner}_8$ ($p$=0)                                  & 37.50                             & 77.41 / 78.70                            & 71.57 / 82.42                          & 84.49                             & 86.19 / 81.81                          & 57.33                            & 89.29                              & 80.37 / 80.44                        & 43.94                              & 73.51                                                   \\
$\text{Learner}_8$ ($p$=2)                                   & 47.72                             & 81.39 / 81.76                            & 83.28 / 88.10                          & 87.94                             & 89.26 / 85.69                          & 59.78                            & 90.46                              & 86.22 / 85.92                        & 33.80                              & 78.34                                                   \\
$\text{Learner}_{64}$ ($p$=0)                                & 44.56                             & 80.67 / 81.14                            & 80.15 / 86.11                          & 87.01                             & 88.28 / 84.42                          & 58.48                            & 89.89                              & 84.48 / 84.27                        & 45.35                              & 76.84                                                   \\
$\text{Learner}_{64}$ ($p$=2)                                & 50.25                             & 81.69 / 82.29                            & 84.66 / 89.23                          & 88.53                             & 89.55 / 86.07                          & 60.07                            & 90.46                              & 86.68 / 86.35                        & 42.54                              & 79.07                                                   \\
$\text{Learner}_{256}$ ($p$=0)                                & 48.79                             & 81.9082.25                             & 84.66 / 89.29                          & 88.64                             & 89.66 / 86.27                          & 60.51                            & 90.41                              & 86.39 / 86.12                        & 43.10                              & 78.95                                                   \\
$\text{Learner}_{256}$ ($p$=2)                                 & 51.18                             & \textbf{82.35 / 82.49}                   & 84.71 / 89.41                          & \textbf{89.12}                    & 90.00 / 86.68                          & 59.86                            & 90.02                              & \textbf{86.95 / 86.60}               & 47.89                              & 79.34                                                   \\ \bottomrule
\end{tabular}
\vskip -0.1in
\end{table*}

Learners address the question of effective training by exploiting pre-trained parameters for their quick convergence. Adapters, as was shown in Figure \ref{figure_cola_convergence}, do not converge quickly on compressed models. This is a function of both their use of newly-initialized parameters as well as a small number of trained parameters. FAR, on the other hand, converged quickly, but trained a greater proportion of parameters. To gain the benefits of both quick convergence and a small number of trained parameters, we rethink the priming technique from FAR \cite{FAR}. Priming the Learner modules instead takes the form of training the entire multi-head attention alongside the projection matrices for a number of initial epochs, $p$, while the rest of the model is frozen. In this way, we start off by training the relatively small multi-head attention linear layers and the Learner projections, switching to training only the projections in the latter epochs. Training alongside pre-trained parameters during priming ensures that the Learner modules have a high rate of convergence across the entirety of fine-tuning. 

To show that Learners may be collapsed and added to the linear layer weights, we present a breakdown of the architecture. The projection matrices $\mathbf{P}_1 \in \mathbb{R}^{l \times d}$ and $\mathbf{P}_2 \in \mathbb{R}^{h \times l}$, where $l$ is the learner hidden size, $d$ is the input size of the linear layer, and $h$ is the output size of the linear layer, can be thought of as an update matrix applied after backpropagation to a linear layer weight matrix, $\mathbf{W}$. The weight matrix after the learners are collapsed into it is $\widetilde{\mathbf{W}}$. This ``collapse'' can similarly be imagined as an identity matrix, $\mathbf{I}_d$, being input with $\widetilde{\mathbf{W}}$ as the combination of the weights. 

\begin{equation}
    \mathbf{I}_d \widetilde{\mathbf{W}}^\intercal = \mathbf{I}_d \mathbf{W}^\intercal + \mathbf{I}_d \mathbf{P}_1^\intercal \mathbf{P}_2^\intercal
\end{equation}

This may be further reduced:
\begin{equation}
    \widetilde{\mathbf{W}} = \mathbf{W} + \mathbf{P}_2 \mathbf{P}_1
\end{equation}

Thus, after fine-tuning, the Learner module may be added to the weight matrix. Henceforth, Learners are denoted as $\text{Learner}_l$ ($p$=x) to specify their hidden size $l$ and the number of priming epochs $x$. Altogether, Learner modules occupy a middle ground between FAR, adapters, and \textsc{Freeze FFNs}. Learners train significantly fewer parameters than FAR and \textsc{Freeze FFNs} while maintaining the benefits of quick convergence.

\section{Experiments}

\paragraph{Fine-Tuning Details}
Learners and their counterparts are fine-tuned for five epochs to simulate realistic device constraints on fine-tuning (i.e., in time, or computational resource usage). The datasets of the GLUE benchmark are used for fine-tuning and an average GLUE score is reported for each tested method \cite{wang2018glue}. Following standard practice, we have reported WNLI but have omitted it from the validation GLUE score. 

All training is done on DistilBERT (Baseline). Learner modules are initialized to zero\footnote{Similarly to adapters, we want to preserve the linear layer function prior to training, i.e., $\widetilde{\mathbf{W}} = \mathbf{W}$.}. We used an adapter hidden size of 256 to give the best chance at convergence, since we found that larger hidden sizes for adapters produced better results. Parallel adapters used a hidden size of 512, a scaling factor of 4, with the adapter parallel to the FFN. Adapter-based methods had all parameters other than layer normalization frozen. FAR was re-implemented and run with a priming percentage of 1\% and retention percentage of 10\%, denoted as $\text{FAR}_{10}$. In all experiments, the classifier layers appended to the model for fine-tuning were trained. We used a batch size of 16 with a learning rate of $2e^{-5}$ and a linear learning rate scheduler. As standard, AdamW is used as the optimizer. Each experiment is run for five epochs, five times, each with a new random seed. Results are reported as the average of those runs. 

\paragraph{Analysis of Learners}
From the results in Table \ref{table_glue_results}, it is clear that Learner modules are capable of fine-tuning DistilBERT without losing points on the GLUE score. Learners enable a relatively small trained parameterization of 5.96 million parameters, which is slightly more than adapters (5.34M), and $7 \times$ less than FAR (41.45M). Learners still, however, maintain a high convergence rate and low training time despite their small trained parameterization\footnote{The training losses of all GLUE tasks except WNLI are illustrated in the appendix. Similarly to Figure \ref{figure_cola_convergence}, learners outpace adapters in convergence rate for all tasks.}. Notably, while adapters fine-tune faster than learners (33.01 minutes vs. 40.50 minutes respectively, c.f., Table \ref{table_paramerization}), the slow convergence rate of adapters means that in five epochs, Learner modules achieve a higher GLUE score. For example, comparing sequential adapters, $\text{Learner}_{64}$ ($p$=0) and $\text{Learner}_{64}$ ($p$=2), we see GLUE scores of 76.10, 76.84, and 79.07 respectively. This may be due to the ability of primed Learner modules to maintain a high convergence rate even after priming. For example, in Figure \ref{figure_cola_convergence}, and the figures in the appendix, it is clear that even after their 2 priming epochs, Learner modules are still faster than adapters at converging, especially within the 5 epoch window of these experiments. On the other hand, FAR and \textsc{Freeze FFNs} perform similarly to $\text{Learner}_{64}$ ($p$=2) in GLUE score and rate of convergence. However, $\text{Learner}_{64}$ ($p$=2) consumes much less memory than FAR in terms of peak memory (1.017 GB vs. 0.793 GB), due to FAR's complex memory operations and higher trained parameter count. Learners also fine-tune $2.5\times$ fewer parameters than \textsc{Freeze FFNs}. Finally, BitFit drastically underperforms all other related works, regardless of its efficiency.

It is evident in Table \ref{table_glue_results} that larger Learner modules produce better results with GLUE scores consistently increasing from $\text{Learner}_{8}$ to $\text{Learner}_{256}$. Larger Learner modules are expected to produce better results due to their larger trained parameterization. However, by adding just two priming epochs, the benefits of faster convergence are realized. In each experiment, those Learner modules with priming outperform those without. $\text{Learner}_{64} (p=2)$ even outperforms the much larger $\text{Learner}_{256} (p=0)$, with GLUE scores of 79.07 and 78.95 respectively. The double bind of efficient fine-tuning, effective and efficient training by a reduction in trained parameters while maintaining quick convergence, has been realized by $\text{Learner}_{64}$ ($p$=2). Furthermore, priming seems to be the key to unlocking efficient training, despite it's greater resource usage in the initial epochs of fine-tuning. Learner modules with priming are thus successful at making a temporary trade-off of efficiency for fine-tuning efficacy. 

\section{Conclusion}
We propose Learner modules, a more efficient and effective alternative to works in parameter-efficient and efficient fine-tuning. Learner modules are shown to train efficiently while also achieving high metric scores in all tested tasks. Future work should explore alternative methods to increase convergence speed. Some alternatives may include distillation or pre-fine-tuning on an out-of-distribution task similar to the target task. Alternative models such as MobileBERT should also be considered due to their even more limited parameterization. 

\section*{Acknowledgements}
We would like to thank Huawei Canada for supporting this work, and Compute Canada for providing computational infrastructure including GPUs and compute servers. We would also like to thank the reviewers for their thoughtful comments and critiques. 




\bibliography{bib}
\bibliographystyle{icml2022}

\newpage
\appendix
\onecolumn
\section{Training Loss Curves for GLUE Tasks}

CoLA training loss curves for related works are shown in the main paper in Figure \ref{figure_cola_convergence}. This appendix presents the training loss curves of the remaining GLUE tasks. These graphs in their totality demonstrate that learner modules have a faster convergence rate than adapters and BitFit while also converging to a lower loss value in fewer steps. Large datasets such as MNLI are run for 5 epochs while smaller datasets are run for 20 epochs. This is to ensure smaller datasets could fully converge. Note that WNLI is not presented here because the dataset is so small as to not produce a useful loss curve (only one loss evaluation in 20 epochs). 

It is noted that after 5 epochs, across all training runs, learners have a lower loss value than adapters. It is only after many more epochs that parallel adapters catch up, e.g., with MRPC, RTE, and STS-B. 

\begin{figure}[!htbp]
  \centering
  \begin{minipage}[b]{0.45\textwidth}
    \includegraphics[width=\textwidth]{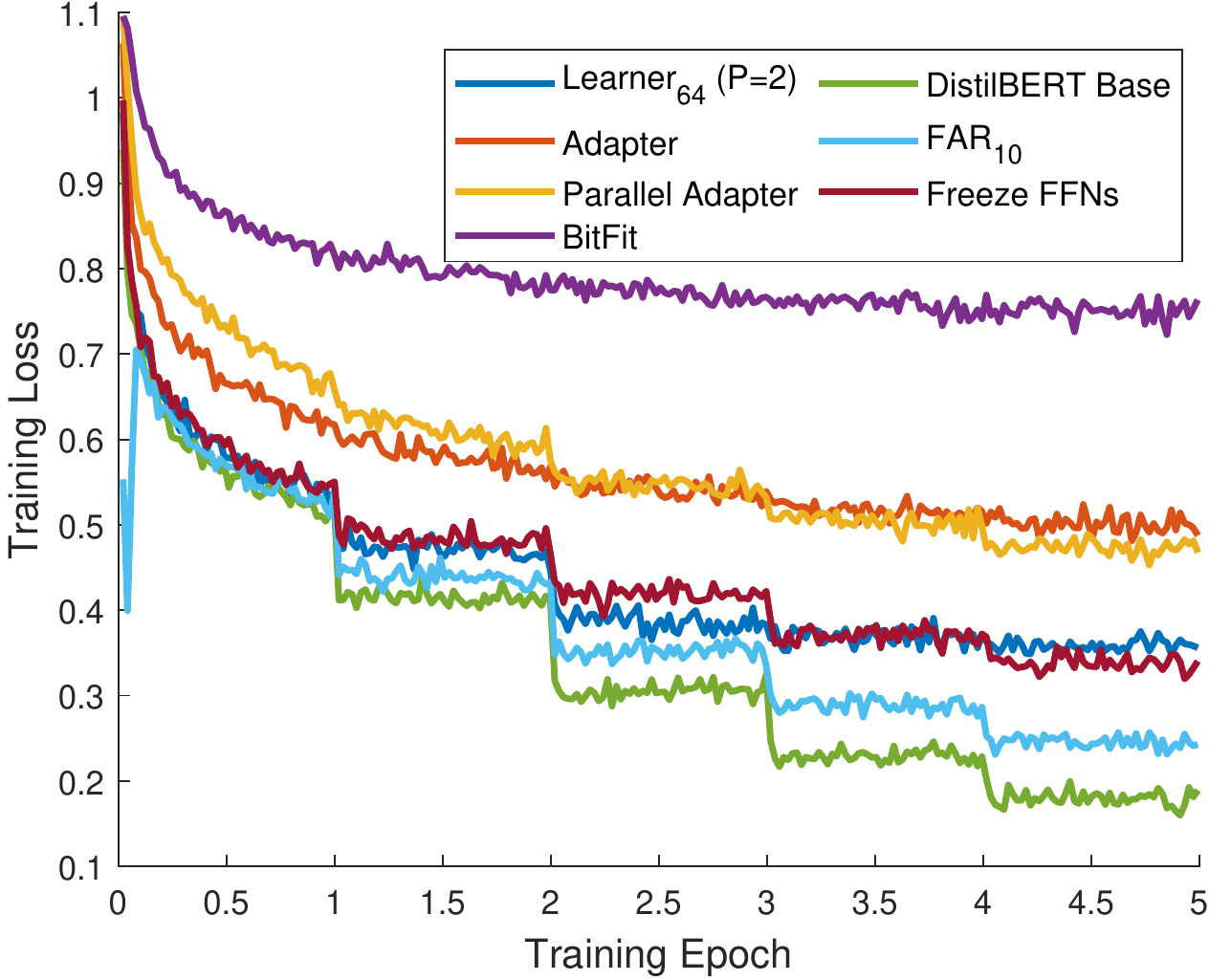}
    \caption{MNLI training loss curve on 5 epochs.}
  \end{minipage}
  \hfill
  \begin{minipage}[b]{0.45\textwidth}
    \includegraphics[width=\textwidth]{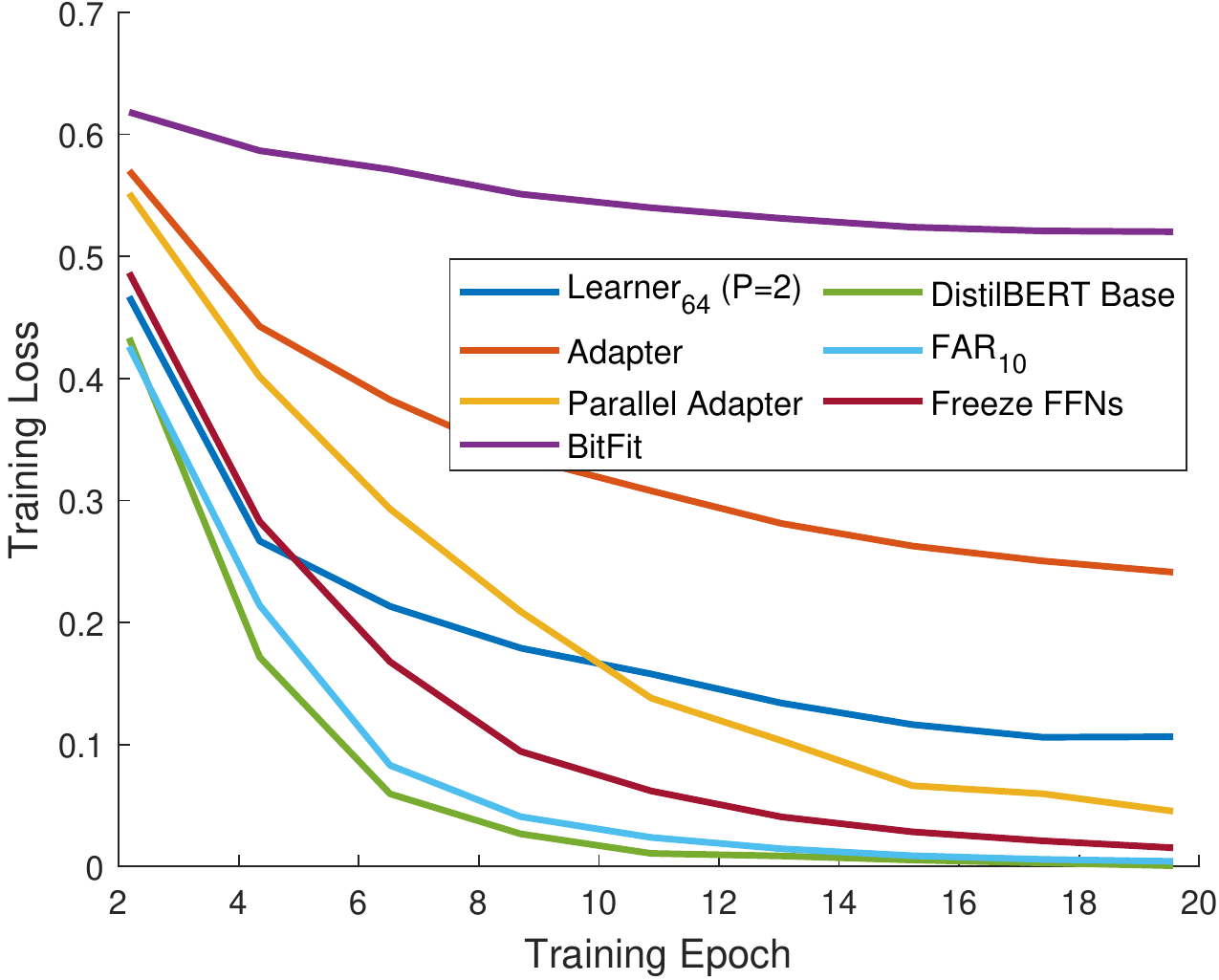}
    \caption{MRPC training loss curve on 20 epochs.}
  \end{minipage}
\end{figure}

\begin{figure}[!htbp]
  \centering
  \begin{minipage}[b]{0.45\textwidth}
    \includegraphics[width=\textwidth]{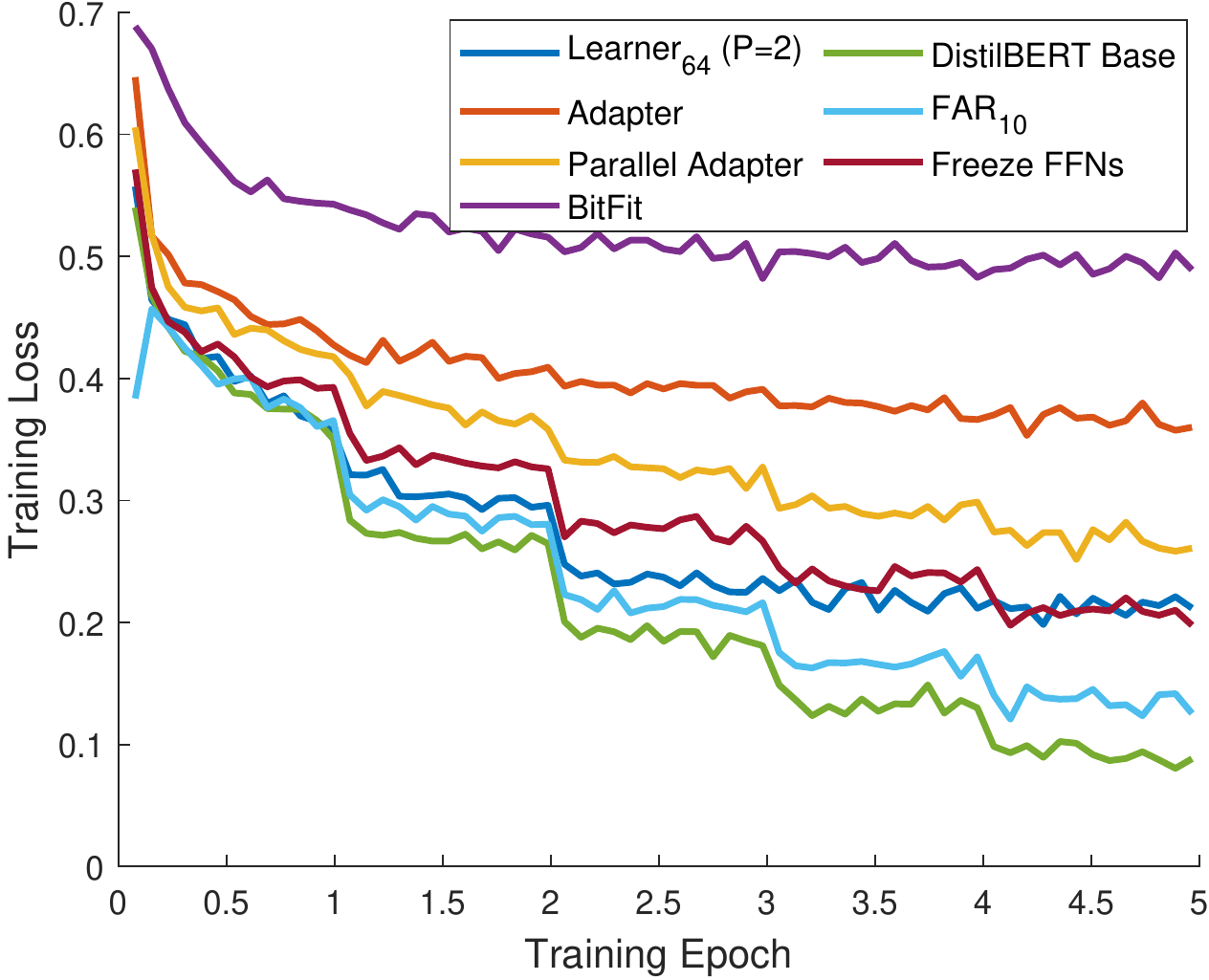}
    \caption{QNLI training loss curve on 5 epochs.}
  \end{minipage}
  \hfill
  \begin{minipage}[b]{0.45\textwidth}
    \includegraphics[width=\textwidth]{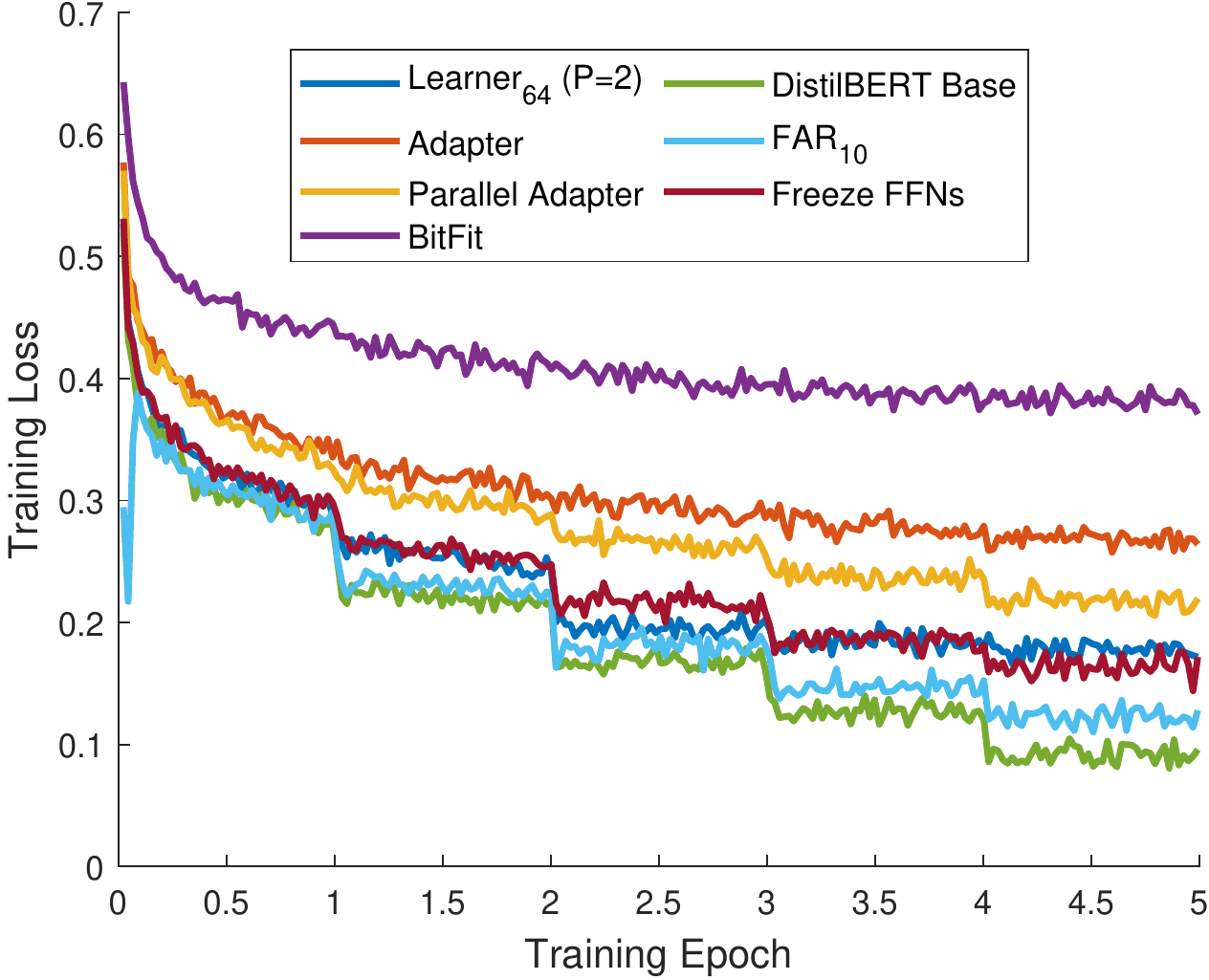}
    \caption{QQP training loss curve on 5 epochs.}
  \end{minipage}
\end{figure}

\begin{figure}[!htbp]
  \centering
  \begin{minipage}[b]{0.45\textwidth}
    \includegraphics[width=\textwidth]{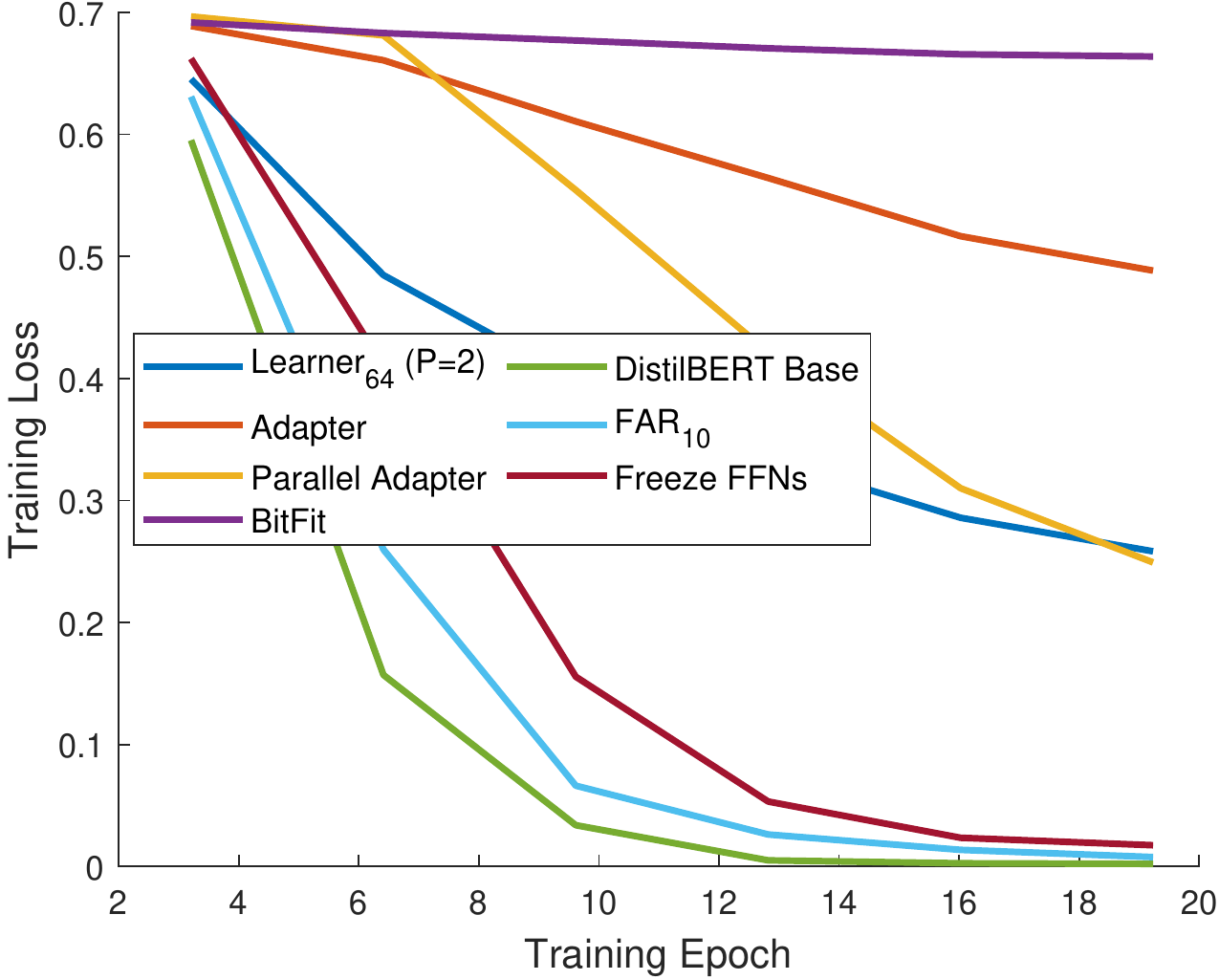}
    \caption{RTE training loss curve on 20 epochs.}
  \end{minipage}
  \hfill
  \begin{minipage}[b]{0.45\textwidth}
    \includegraphics[width=\textwidth]{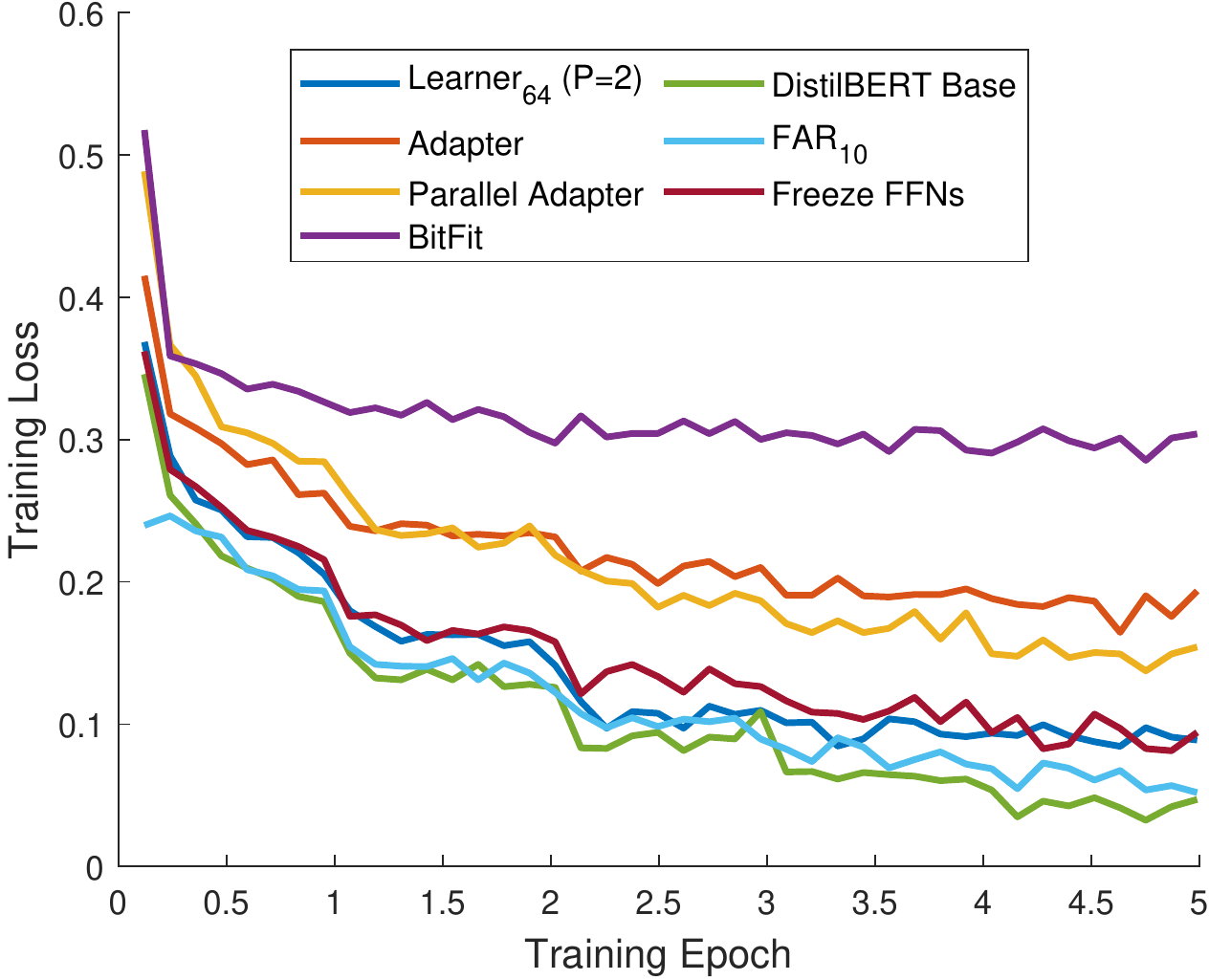}
    \caption{SST-2 training loss curve on 5 epochs.}
  \end{minipage}
\end{figure}

\begin{figure}[ht]
\vskip 0.2in
\begin{center}
\centerline{\includegraphics[width=0.45\textwidth]{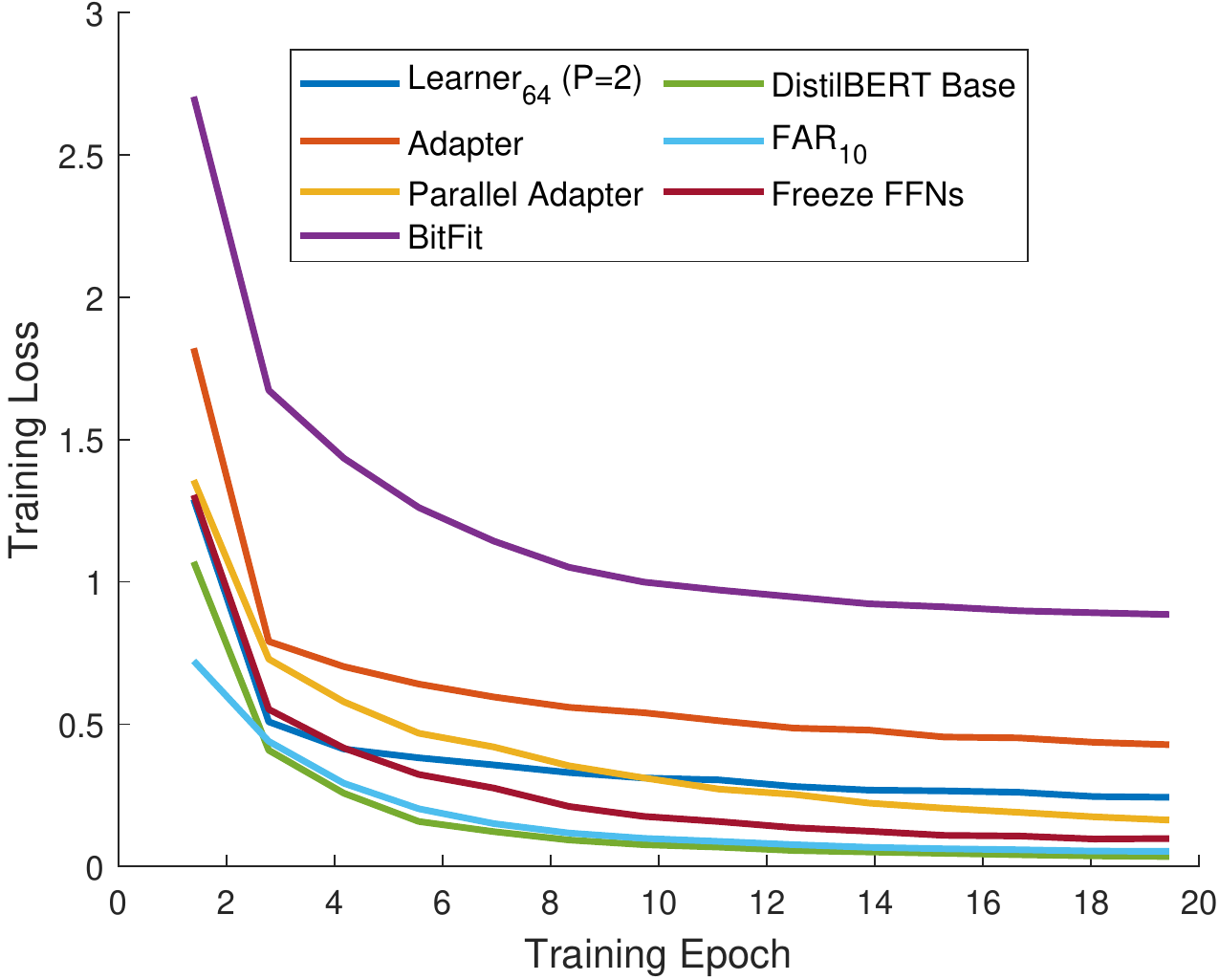}}
\caption{STS-B training loss curve on 20 epochs.}
\end{center}
\vskip -0.2in
\end{figure}


\end{document}